# AN EXHAUSTIVE STUDY OF THE WORKSPACE TOPOLOGIES OF ALL 3R ORTHOGONAL MANIPULATORS WITH GEOMETRIC SIMPLIFICATIONS

Mazen Zein, Philippe Wenger and Damien Chablat
IRCCyN, Institut de Recherche en Communications et Cybernetique de Nantes
1, rue de la Noe – BP 92101 44321 Nantes Cedex 3 France
e-mail: Mazen.Zein@irccyn.ec-nantes.fr

ABSTRACT- This paper proposes a classification of three-revolute orthogonal manipulators that have at least one of their DH parameters equal to zero. This classification is based on the topology of their workspace. The workspace is characterized in a half-cross section by the singular curves. The workspace topology is defined by the number of cusps and nodes that appear on these singular curves. The manipulators are classified into different types with similar kinematic properties. Each type is evaluated according to interesting kinematic properties such as, whether the workspace is fully reachable with four inverse kinematic solutions or not, the existence of voids, and the feasibility of continuous trajectories in the workspace. It is found that several orthogonal manipulators have a "well-connected" workspace, that is, their workspace is fully accessible with four inverse kinematic solutions and any continuous trajectory is feasible. This result is of interest for the design of alternative manipulator geometries.

## 1 Introduction

The workspace of general 3R manipulators has been widely studied in the past (see, for instance, [1-5]). The determination of the workspace boundaries, the size and shape of the workspace, the existence of holes and voids, the accessibility inside the workspace (i.e. the number of inverse kinematic solutions in the workspace), are some of the main features that have been explored. Today, most industrial manipulators are of the PUMA type; they have a vertical revolute joint followed by two parallel joints and a spherical wrist. Another interesting category of serial manipulators exists, which have any two consecutive joint axes orthogonal. We call these manipulators *orthogonal manipulators*. Instances of orthogonal manipulators are the IRB 6400C launched by ABB-Robotics in 1998 and the DIESTRO manipulator built at McGill University.

Unlike PUMA type manipulators, orthogonal manipulators may have many different kinematic properties according to their links and joint offsets lengths, so it is interesting to classify them. Orthogonal 3R manipulators may be binary (only two inverse kinematic solutions) or quaternary (four inverse kinematic solutions), they may have voids or no voids in their workspace, they may be cuspidal or noncuspidal. A cuspidal manipulator is one that can change posture without meeting a singularity [6, 7]. Several conditions for a manipulator to be noncuspidal were provided in [8, 9] and a general necessary and sufficient condition for a 3-DOF manipulator to be cuspidal was established in [10], which is the existence of at least one point, called cusp point, in the workspace where the inverse kinematics admits three equal solutions.

In [11], the authors established a categorization of all generic 3R manipulators based on the homotopy classes of their singularities in the joint space. More recently, [12] attempted the classification of 3R orthogonal manipulators with no offset on their last joint. Three surfaces were found to divide the manipulator parameter space into cells with equal number of cusp points. The equations of these surfaces were derived as polynomials in the DH-parameters using Groebner Bases. The work of [12] was completed in [13] to take into account additional features in the classification, such as genericity. The authors of [14] established a classification of 3R orthogonal manipulators with no offset on their last joint, based on the work of [13], according to the number of cusps and node points. The parameter space was divided into nine cells where the manipulators have the same number of cusps and nodes in their workspace.

About ten remaining families of 3R manipulators with at least one of their DH parameters equal to zero have not been classified yet. Because industrial manipulators usually have at least one parameter equal to zero, it is of practical interest to classify them.

The purpose of this paper is to classify a family of 3R orthogonal manipulators that have at least one of their DH parameters equal to zero. This classification is based on the topology of their workspace. The workspace is characterized in a half-cross section by the singular curves. The workspace topology is defined by the number of cusps and nodes that appear on these singular curves. These singular points are interesting features for characterizing the 4-solution regions and the voids in the workspace. The manipulators are classified into different types with similar kinematic properties. Each type is evaluated according to interesting kinematic features such as, whether the workspace is fully reachable with four inverse kinematic solutions or not, the existence of voids, and the feasibility of continuous trajectories in the workspace. It is found that several orthogonal manipulators have a "well-connected" workspace [15], that is, the workspace is fully accessible with four inverse kinematic solutions and any continuous trajectory is feasible throughout the workspace. This result is of interest for the design of alternative manipulator geometries.

Next section of this article presents the families of manipulators under study and recalls some preliminary results. The classifications are established in section 3. Section 4 analyzes the resulting classification and several interesting manipulator geometries are pointed out. Section 5 concludes this paper.

## 2    Preliminaries

### 2.1    Orthogonal manipulators

The manipulators studied, referred to as *orthogonal manipulators*, are positioning manipulators with three revolute joints in which the two pairs of adjacent joint axes are orthogonal. The length parameters are $d_2$, $d_3$, $r_2$, $r_3 \geq 0$ and $d_4 > 0$ ($d_4$ cannot be equal to zero because the resulting manipulator would be always singular). The angle parameters $\alpha_2$ and $\alpha_3$ are set to $-90°$ and $90°$, respectively. The three joint variables are referred to as $\theta_1$, $\theta_2$ and $\theta_3$, respectively. They will be assumed unlimited in this study. The position of the end-tip is defined by the three Cartesian coordinates *x*, *y* and *z* of the operation point P with respect to a reference frame (O, **X**, **Y**, **Z**) attached to the manipulator base. Figure 1 shows the architecture of the manipulators under study in the home configuration defined by $\theta_1=\theta_2=\theta_3=0$.

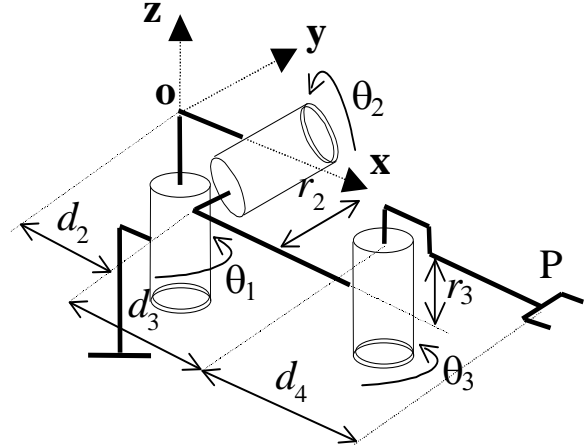

Fig. 1: A 3R orthogonal manipulator.

## 2.2 Manipulators with geometric simplifications

In the classification provided in [13,14], the only particular case considered was $r_3=0$ with the remaining parameters being all different from zero. Here, we will treat all the other possible combinations of manipulators with at least one DH parameter equal to zero. The different combinations are depicted by the tree shown in Fig. 2, which yields ten families to analyze. Recall that we cannot have $d_4=0$.

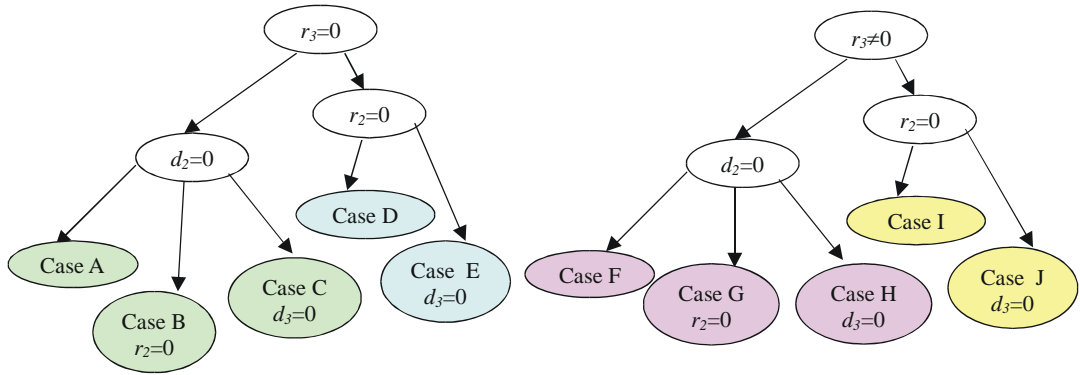

Fig. 2: The ten families of manipulators to analyze with at least one parameter equal to zero.

## 2.3 Singular curves, cusp points and node points

The singularities of a manipulator play an important role in its global kinematic properties [7, 8]. The singularities of a general 3R manipulator can be determined with the determinant of the Jacobian matrix det(**J**). For an orthogonal manipulator, det(**J**) takes on the following form [20]

$$\det(\mathbf{J}) = d_4 \left\{ (d_3 + d_4 c_3)\left[(d_2 + r_3 s_2)s_3 + (d_3 s_3 - r_2 c_3)c_2\right] + r_3(r_2 + d_4 s_3)s_2 c_3 \right\} \quad (1)$$

where $c_2=\cos(\theta_2)$ and $s_2=\sin(\theta_2)$, $c_3=\cos(\theta_3)$ and $s_3=\sin(\theta_3)$.

Since the singularities are independent of $\theta_1$, the contour plot of $\det(\mathbf{J})=0$ can be displayed in $-\pi \leq \theta_2 < \pi, -\pi \leq \theta_3 < \pi$ where they form a set of close curves. These curves divide the joint space into singularity-free domains called *aspects* [21]. The singularities can also be plotted in the workspace by searching for the points where the inverse kinematics has double roots [2], or using $\det(\mathbf{J})=0$ and the direct kinematics. Because of its symmetry about the first joint axis, the workspace may be analyzed by its half cross-section defined by ($\rho = \sqrt{x^2 + y^2}$, $z$). When plotted in this section, the singularities define the internal and external boundaries curves of the workspace cross-section. Figure 3 illustrates the singularity curves for a 3R orthogonal manipulator with no offset along its last joint axis ($r_3=0$). When $r_3=0$, $\det(\mathbf{J})$ takes on the factored form $\det(\mathbf{J}) = (d_3 + d_4 c_3)[d_2 s_3 + (s_3 d_3 - c_3 r_2)c_2]$. The first factor defines two horizontal lines in the joint space (assuming $d_3 \leq d_4$, which is the case for the manipulator in Fig. 3). The singular line defined by $\theta_3=+\arccos(-d_3/d_4)$ maps onto one singular point in the workspace cross-section, which is located at the self-intersection of the internal singular boundary. The remaining singular line $\theta_3=-\arccos(-d_3/d_4)$ maps onto an isolated singular point in the workspace. One of the two singular curves defines the external boundary of the workspace and the other one defines the internal boundary.

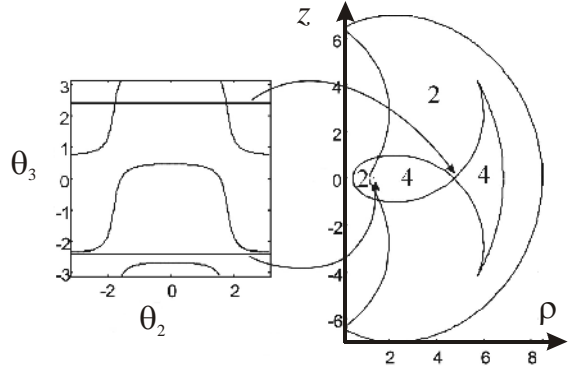

Fig. 3: The singular curves of a 3R orthogonal manipulator with $d_2=1$, $d_3=3$, $d_4=4$, $r_2=2$ and $r_3=0$.

An interesting classification criterion is the topology of the singular curves in the cross section of the workspace. A way of defining the topology of these curves is to enumerate their singular points, the cusp points and the node points [14, 16].

A cusp point is one where the inverse kinematics admits three equal solutions [10]. A 3-DOF positioning manipulator can change its posture without meeting a singularity only by encountering a cusp point [10].

At a node point, the inverse kinematics admits two pairs of coincident inverse kinematic solutions [13]. Two singular curves intersect at a node point. For the manipulator shown in Fig. 3, the workspace features two cusps and three nodes. One of the nodes is at the self-intersection of the internal boundary and the other two nodes are at the intersection of the internal boundary with the external boundary. These cusps and nodes define two regions with two inverse kinematic solutions and two regions with four inverse kinematic solutions.

The existence of cusps and nodes can be determined from the fourth-degree inverse kinematic polynomial of 3R manipulators, which can be written as a polynomial $P(t)$ in $t = \tan(\theta_3/2)$ whose coefficients are function of the design parameters $d_2$, $d_3$, $d_4$, $r_2$ and $r_3$ and of the variables $R = x^2 + y^2$ and $Z = z^2$ (see [2] for instance).

Node points appear when $P(t)$ has two pairs of equal roots. A condition for $P(t)$ to have two pairs of equal roots is

$$\begin{cases} P(t) = 0 \\ \dfrac{\partial P}{\partial t}(t) = 0 \\ \Delta\left(\dfrac{\partial P}{\partial t}\right) = 0 \end{cases} \quad (2)$$

where $\Delta$ means the polynomial discriminant. Cusp points appear when $P(t)$ has triple roots. A condition for $P(t)$ to have triple roots is

$$\begin{cases} P(t) = 0 \\ \dfrac{\partial P}{\partial t}(t) = 0 \\ \dfrac{\partial^2 P}{\partial t^2}(t) = 0 \end{cases} \quad (3)$$

2.4   Separating surfaces in the parameter space and classification of manipulator geometries

The main results of [20] are briefly reviewed in this section. The parameter space of orthogonal manipulators was divided by *separating surfaces*, where the number of cusps or nodes changes. In other words, these surfaces are associated with transition manipulators and define several domains in the parameter space where all manipulators have the same number of cusps and nodes and the same pattern of singular curves. Consequently, moreover, all manipulators in a given domain of the parameter space will have the same number of 4-solution regions and voids in their workspace. For example, all manipulators belonging to the same domain as the one shown in Fig. 3 will have no voids in their workspace and for all of them the internal boundary curve will look like a "fish" with the "tail" never cutting the external workspace boundary and the "head" always intersecting this boundary, thus defining two 4-solution regions and two 2-solution regions. Note that since the classification is based on the topology only, the size of these regions will depend on the value of the parameters.

The separating surfaces were determined in two steps. The first step is a formal mathematical approach that consists in deriving the condition under which the number of solutions to (2) or (3) changes. This condition was solved using Groebner bases [12]. The second step is a numerical verification that aims at eliminating the spurious solutions. Since the manipulators that we want to classify here are such that $d_2=0$ or $d_3=0$ or $r_2=0$, we know that their inverse kinematic polynomial can be written as a quadratic and, thus, it cannot admit triple roots [2, 17]. Consequently, we know that these manipulators will be always non-cuspidal. Thus, the only separating surfaces of interest are those associated with a change in the number of nodes.

The classification proposed in this paper is not a direct extension of the one that was conducted in [20]. First, in [20] all parameters were normalized by $d_2$ to reduce the dimension of the problem, thus assuming $d_2=1$. Second, the equations of the separating surfaces found in [20] may take degenerate forms when the value of one or more parameter is zeroed. Thus, a careful case by case analysis is required, which was not conducted before.

The separating surfaces associated with a change in the number of nodes found in [20] are recalled below. For manipulators with $r_3=0$, the separating surfaces were found to be defined by the following three equations (for $d_2=1$)

$$\left.\begin{aligned}(E1): d_4 &= \frac{1}{2}(a-b) \\ (E2): d_4 &= d_3 \\ (E3): d_4 &= \frac{1}{2}(a+b)\end{aligned}\right\} \quad (4)$$

with:

$$a = \sqrt{(d_3+1)^2 + r_2^2} \text{ and } b = \sqrt{(d_3-1)^2 + r_2^2} \quad (5)$$

For an arbitrary $d_2$, $\alpha$ and $\beta$ take the following form:

$$a = \sqrt{(d_3+d_2)^2 + r_2^2} \text{ and } b = \sqrt{(d_3-d_2)^2 + r_2^2} \quad (6)$$

For manipulators with $r_3 \neq 0$, the separating surfaces associated with a change in the number of nodes found in [20] are (for $d_2=1$)

$$\left.\begin{aligned} d_4 &= \sqrt{\frac{1}{2}\left(1 + r_2^2 + d_3^2\left(1-\left(\frac{r_3}{r_2}\right)^2\right)\right) + \left(\frac{r_3}{r_2}\right)^2 - r_3^2 - \alpha\beta\left|\left(\frac{r_3}{r_2}\right)^2 - 1\right|} \\ d_4 &= \sqrt{\frac{1}{2}\left(1 + r_2^2 + d_3^2\left(1-\left(\frac{r_3}{r_2}\right)^2\right)\right) + \left(\frac{r_3}{r_2}\right)^2 - r_3^2 + \alpha\beta\left|\left(\frac{r_3}{r_2}\right)^2 - 1\right|} \end{aligned}\right\} \quad (7)$$

$$(\Sigma 1): d_4 = \sqrt{d_3^2 + r_2^2} \quad (8)$$

with $\alpha$ and $\beta$ defined as in (5).

Equations (7) were derived for $r_2 \neq 0$ and $d_2=1$. For an arbitrary $d_2$, Eqs (7) take the following form:

$$d_4 = \sqrt{\frac{1}{2}\left(d_2^2 + r_2^2 + d_3^2\left(1 - \left(\frac{r_3}{r_2}\right)^2\right) + d_2^2\left(\frac{r_3}{r_2}\right)^2 - r_3^2 - \alpha\beta\left|\left(\frac{r_3}{r_2}\right)^2 - 1\right|\right)}$$

$$d_4 = \sqrt{\frac{1}{2}\left(d_2^2 + r_2^2 + d_3^2\left(1 - \left(\frac{r_3}{r_2}\right)^2\right) + d_2^2\left(\frac{r_3}{r_2}\right)^2 - r_3^2 + \alpha\beta\left|\left(\frac{r_3}{r_2}\right)^2 - 1\right|\right)}$$

(9)

where $\alpha$ and $\beta$ are defined as in Eqs. (6).

In fact, the above two equations define the two (positive) roots of the following general equation:

$$(\Sigma 2): \begin{cases} d_4^4 r_2^2 + \left(-r_2^4 - d_3^2 r_2^2 + r_3^2 r_2^2 - d_2^2 r_2^2 + d_3^2 r_3^2 - d_2^2 r_3^2\right) d_4^2 \\ + d_2^2 d_3^2 r_2^2 - d_2^2 r_3^4 + d_2^4 r_3^2 + d_2^2 r_2^2 r_3^2 - d_2^2 d_3^2 r_3^2 = 0 \end{cases}$$

(10)

In effect, solving Eq. (10) for $d_4^2$ for strictly positive values of $r_2$ yields Eqs. (9) with $\alpha$ and $\beta$ defined as in (6). Because we may have $r_2=0$ in the following classification, Eq. (10) will be used rather than Eqs. (9).

In section 3, a case by case analysis of Eqs. (4), (8) and (10) will be conducted for the ten families of manipulators shown in Fig. 2.

## 2.5 Feasibility of continuous trajectories in the workspace

An important kinematic feature is the feasibility of continuous trajectories within the workspace. This feature is important for process tasks such as welding or cutting. For non-cuspidal manipulators, the regions of feasible continuous trajectories in the workspace are the images of the aspects under the kinematic map [18,19,21]. This is because this map being one-to-one from each aspect onto its image in the workspace, the preimage of any continuous trajectory of the workspace is a path in the joint space. The workspace of a manipulator is said to be *t-connected* if any continuous trajectory is feasible, which arises when one region of feasible continuous trajectories is found to be coincident with the whole workspace [18]. In [19], a general algorithm was proposed to plot the regions of feasible continuous trajectories in a cross section of the workspace. This algorithm works for both cuspidal and non-cuspidal manipulators.

When the workspace is t-connected and composed of only one 4-solution region, it is said to be well-connected [15]. This is an interesting feature, which arises in Puma manipulators with equal link lengths [15]. It can be shown that if a given manipulator has a t-connected (resp. well-connected) workspace, then all manipulators belonging to the same domain of the parameter space will have a t-connected (resp. well-connected) workspace as well [20]. In the following classification, therefore, it will be sufficient to analyze the t-connectivity for one representative manipulator workspace in each domain of the parameter space.

## 3 Classification of the ten families of manipulators

In this section, we classify each family of manipulators shown in Fig. 2. Because nodes on the $z$-axis play no role in the number of 2-solution regions and 4-solution regions or in the t-connectivity analysis [20], they need not be considered. For each family of manipulators,

Eqs. (4), (8) and (10) are rewritten by zeroing the value of one or more parameters. Then, the validity of the resulting equations must be carefully analyzed. A surface will have to be discarded if (*i*) it defines infeasible manipulators (e.g. $d_4$=0), (*ii*) it is associated with the apparition of nodes on the *z*-axis only, or (*iii*) it becomes, for the family of manipulators considered, a spurious solution. Situation (*iii*) arises because the numerical verification that was used in [20] to eliminate the spurious solutions did not consider the particular cases shown in Fig. 2. In the following, for all families of manipulators studied, the validity of each surface was verified numerically by determining the number of nodes of three manipulators defined on the surface and close to it. If the number of nodes was found to be the same for all three manipulators, the equation of the surface was declared spurious and discarded.

In the following classification, cases A through E define manipulators with $r_3$=0 and at least one more parameter (different from $d_4$) equal to zero (see Fig. 2). Thus, all these manipulators will be defined by at most three nonzero parameters. However, because it is always possible to normalize all parameters by one of them, only a 2-dimensional section of the parameter space needs be analyzed. Manipulator families associated with cases G, H and J are defined by only three parameters (see Fig. 2) and only a 2-dimensional section of the parameter space will be shown. For cases F and I, however, the manipulators have four nonzero parameters.

3.1    Case A ($d_2$=0, $r_2$≠0, $d_3$≠0 and $r_3$=0)

The first family of manipulators studied are defined by $d_2$=0 and $r_3$=0. In this case, $d_2$=0 =>$\alpha$=$\beta$ and (*E1*) yields $d_4$=0, which is not possible. Replacing $d_2$=0 and $r_3$=0 in the last two equations of (4) yields

  (*E2*): $d_4 = d_3$     (11)

  (*E3*): $d_4 = \sqrt{d_3^2 + r_2^2}$     (12)

The transition curves (*E2*) and (*E3*) divide the parameter space into three domains collecting manipulators with 0, 2 and 4 node points in their workspaces, respectively. Figure 4 shows, for $r_2$=1, the parameter space with the three domains, the transition curves (*E2*) and (*E3*), the workspace of a manipulator in each domain and on each transition. For each workspace, the 4-solution regions are filled in dark gray and the 2-solutions regions are in light gray.

This family of manipulators is classified into the following three types in Fig. 4:

- **Type A1**: The manipulators of this type have $d_3$>$d_4$. Their workspace admits no voids nor node points, it is composed of two 2-solution regions and one 4-solution region. Their workspace is t-connected.

- **Type A2**: The manipulators of this type have $d_3$<$d_4$<$\sqrt{d_3^2 + r_2^2}$. Their workspace admits two node points and no voids. Their workspace is t-connected.

- **Type A3**: The manipulators of this type have $d_4 > \sqrt{d_3^2 + r_2^2}$. Their workspace admits four node points and no voids. Their workspace is not t-connected.

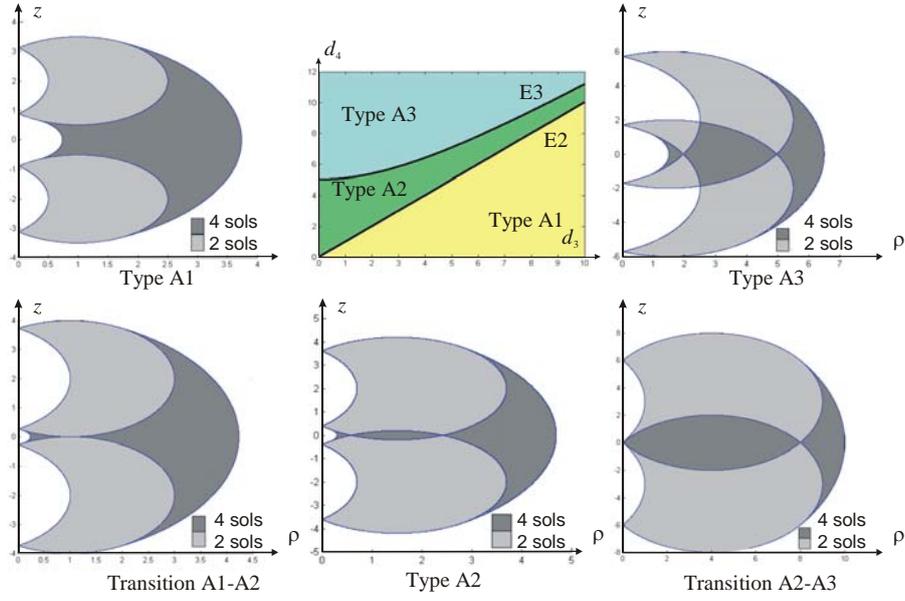

Fig. 4: Parameter space of case A and workspaces of the manipulators having the following parameters: Type A1 ($d_3$=2, $d_4$=1.5 and $r_2$=1); Type A2 ($d_3$=2, $d_4$=2.2 and $r_2$=1.5); Type A3 ($d_3$=2, $d_4$=3 and $r_2$=1). Transition A1-A2 ($d_3$=2, $d_4$=2 and $r_2$=1); Transition A2-A3 ($d_3$=3, $d_4$=5 and $r_2$=4).

### 3.2 Case B ($d_2$=0, $r_2$=0, $d_3 \neq 0$ and $r_3$=0)

These manipulators are such that $d_2$=0, $r_2$=0 and $r_3$=0. Since $d_2$=0, ($E1$) yields $d_4$=0, which is not possible. Moreover, ($E3$) is equivalent to ($E2$). Thus the parameter space is divided, by the transition curve ($E2$), into two domains collecting the manipulators with 0 and 1 node points in their workspaces respectively. Replacing $d_2$=0, $r_2$=0 and $r_3$=0 in (4) yields:

($E2$): $d_3 = d_4$ (13)

Figure 5 below shows the parameter space with two domains, the transition curve ($E1$), the workspace of a manipulator in each domain and on each transition.

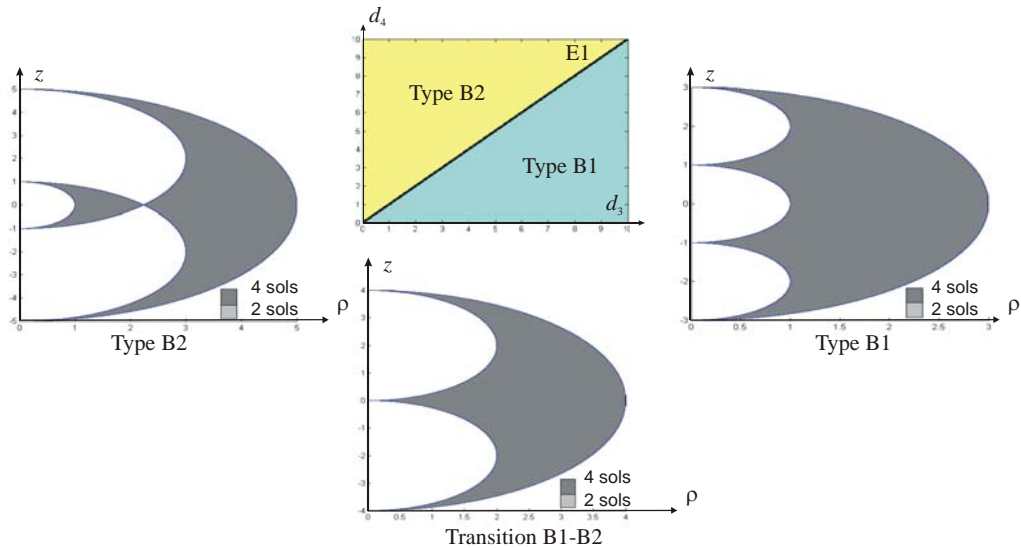

Fig. 5: Parameter space of case B and workspaces of the manipulators having the following parameters: Type B1 ($d_3$=2 and $d_4$=3); Type B1 ($d_3$=2 and $d_4$=1); Transition B1-B2 ($d_3$=2 and $d_4$=2).

This family of manipulators is classified into the following two types:

- **Type B1:** The manipulators of this type have $d_3 > d_4$. Their workspace has no voids and no node points; it is composed of only one 4-solution region. Their workspace is t-connected.

- **Type B2:** The manipulators of this type have $d_4 > d_3$. Their workspace admits one node point and no voids. Their workspace is not t-connected.

### 3.3 Case C ($d_2 = 0$, $r_2 \neq 0$, $d_3 = 0$ and $r_3 = 0$)

These manipulators are such that $d_2 = 0$, $d_3 = 0$ and $r_3 = 0$. Then, ($E1$) and ($E2$) both yield $d_4 = 0$, which is not possible. On the other hand ($E3$) plays no role because it is associated with the apparition of nodes on the $z$-axis. Thus, we have only one type of manipulators, which we call type C. Figure 6 shows the workspace of such manipulators.

- **Type C:** The manipulators of this type do not admit voids nor node points in their workspace, which is composed of only one 4-solution region. Their workspace is t-connected. Thus, it is also well-connected.

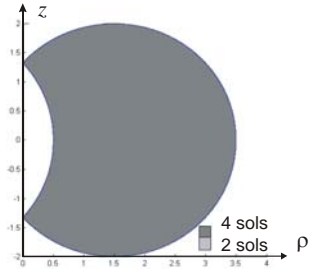

Fig. 6: Workspace of the manipulator having the following parameters $d_4 = 2$ and $r_2 = 1.5$.

### 3.4 Case D ($d_2 \neq 0$, $r_2 = 0$, $d_3 \neq 0$ and $r_3 = 0$)

These manipulators are such that $r_2 = 0$ and $r_3 = 0$. Replacing $r_2 = 0$ and $r_3 = 0$ in (4) yields:

($E1$): ($d_4 = d_2$ and $d_3 > d_2$) or ($d_4 = d_2$ and $d_3 < d_2$) (14)

($E2$): $d_4 = d_3$ (15)

($E3$): ($d_3 = d_2$ and $d_4 < d_2$) or ($d_3 = d_2$ and $d_4 > d_2$) (16)

The transition curves ($E1$), ($E2$) and ($E3$) divide the parameter space into five domains collecting manipulators with 2, 0, 1, 2 and 0 node points in their workspaces, respectively. Figure 7 shows, for $d_2 = 1$, the parameter space with the five domains, the transition curves ($E1$), ($E2$) and ($E3$), the workspace of a manipulator in each domain and on each transition. In this case, we have five types of manipulators, referred to as D1, D2, D3, D4 and D5:

- **Type D1:** The manipulators of this type have $d_4 < d_2 < d_3$. Their workspace admits one void and 2 node points. Their workspace is not t-connected.

- **Type D2:** The manipulators of this type have $d_2 < d_4 < d_3$. Their workspace does not admit voids nor node points. Their workspace is not t-connected.

- **Type D3:** The manipulators of this type have $d_2 < d_3 < d_4$. Their workspace admits one node point and no voids. Their workspace is not t-connected.

- **Type D4:** The manipulators of this type have $d_3 < d_2 < d_4$. Their workspace admits two node points and no voids. Their workspace is not t-connected.

- **Type D5:** The manipulators of this type have $d_3 < d_4 < d_2$. Their workspace does not admit voids nor node points. Their workspace is t-connected.

- **Type D6:** The manipulators of this type have $d_4 < d_3 < d_2$. They are binary, that is, their workspace is composed of only one region, which is reachable with two inverse kinematic solutions. Their workspace admits one void and no node points. Their workspace is t-connected.

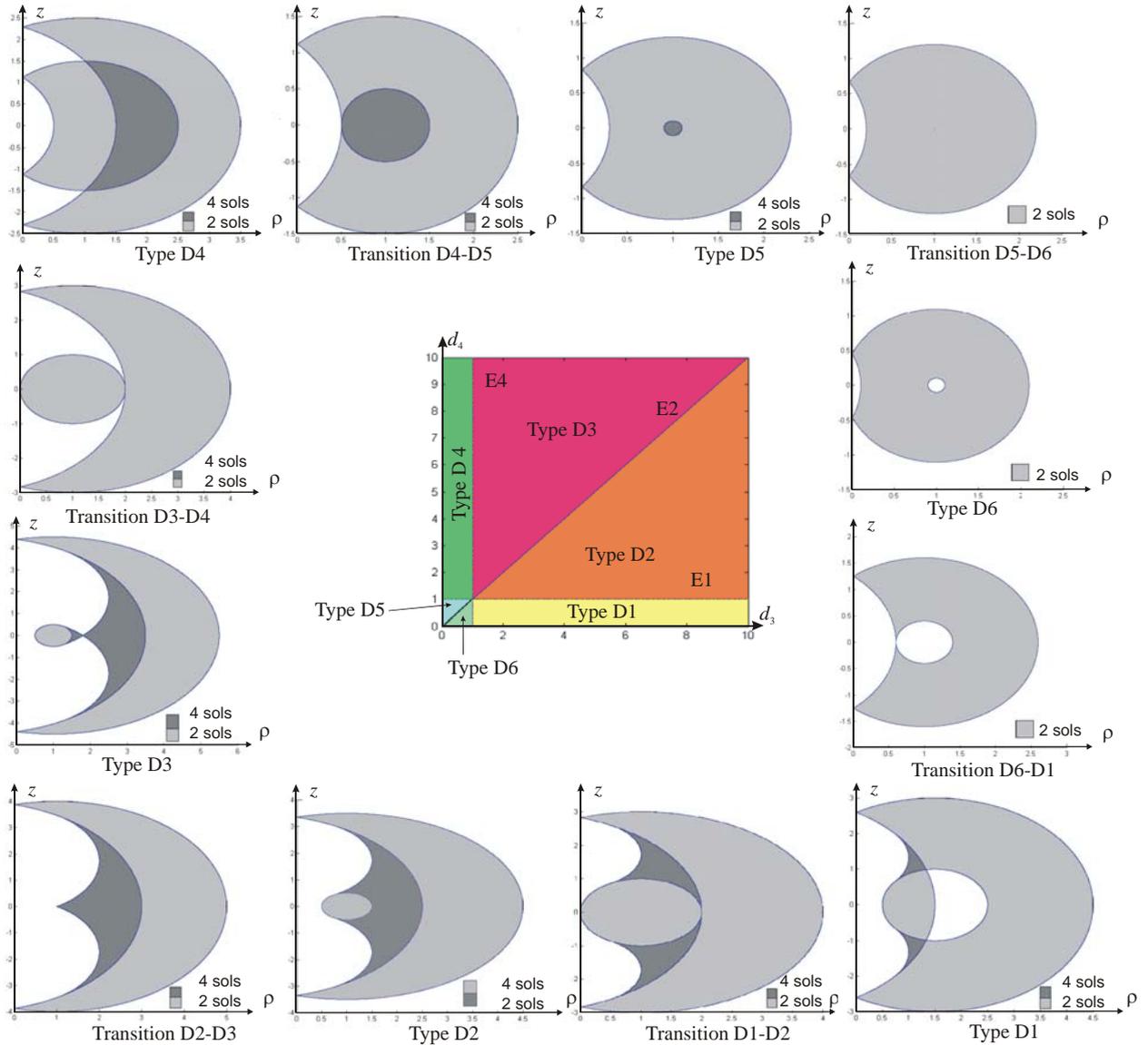

Fig. 7: Parameter space of case D and workspaces of the manipulators having the following parameters: Type D1 ($d_3$=1.4, $d_4$=0.7 and $d_2$=1); Type D2 ($d_3$=2, $d_4$=1.5 and $d_2$=1); Type D3 ($d_3$=2, $d_4$=2.5 and $d_2$=1); Type D4 ($d_3$=0.5, $d_4$=2 and $d_2$=1); Type D5 ($d_3$=0.6, $d_4$=0.7 and $d_2$=1); Type D6 ($d_3$=0.7, $d_4$=0.5 and $d_2$=1); Transition D1-D2 ($d_3$=2, $d_4$=1 and $d_2$=1); Transition D2-D3 ($d_3$=2, $d_4$=2 and $d_2$=1); Transition D3-D4 ($d_3$=1, $d_4$=2 and $d_2$=1); Transition D4-D5 ($d_3$=0.5, $d_4$=1 and $d_2$=1); Transition D5-D6 ($d_3$=0.6, $d_4$=0.6 and $d_2$=1); Transition D6-D1 ($d_3$=1, $d_4$=0.5 and $d_2$=1).

### 3.5 Case E ($d_2 \neq 0$, $r_2 = 0$, $d_3 = 0$ and $r_3 = 0$)

These manipulators are such that $d_3 = 0$, $r_2 = 0$ and $r_3 = 0$. In this case, (*E*1) and (*E*2) both yield $d_4 = 0$ and (*E*3) is associated with the apparition of nodes on the *z*-axis and, thus, is discarded. For this family, thus, we have only one type of manipulators, which we call Type E.

- **Type E** (Fig. 8)**:** The manipulators of this type do not admit voids nor node points in their workspace, which is composed of only one 4-solution region. The workspace is t-connected. Thus, these manipulators have a well-connected workspace.

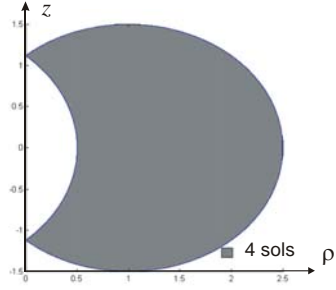

Fig. 8: Workspace of the manipulator having the following parameters $d_4 = 1.5$ and $d_2 = 1$.

### 3.6 Case F ($d_2 = 0$, $r_2 \neq 0$, $d_3 \neq 0$ and $r_3 \neq 0$)

From now on (case F to case J), the parameter $r_3$ is strictly positive. Thus, the equations to inspect are Eqs. (8) and (10). The first family with $r_3 \neq 0$ that we analyze here is defined by $d_2 = 0$. Equation (10) appears to be an extraneous solution for this particular case as we verified numerically that it plays no role. On the other hand, Eq. (8) does play a role and divides the parameter space into two domains where the manipulators have 0 or 2 node points in their workspace, respectively. Figure 9 shows a section of the parameter space for $r_2 = 1$ and an arbitrary value of $r_3$ (the transition is independent of $r_3$ since this parameter does not appear in Eq. (8)). The equation of the transition curve is exactly Eq. (8), which we recall below:

$$(\Sigma 1) \; : \; d_4 = \sqrt{d_3^2 + r_2^2} \qquad (17)$$

For this family, there are two types of manipulators, referred to as F1 and F2, respectively:

- **Type F1:** The manipulators of this type have $d_4 < \sqrt{d_3^2 + r_2^2}$. Their workspace is free of voids and node points. It is composed of two 2-solutions regions and one 4-solution region. Their workspace is t-connected.

- **Type F2:** The manipulators of this type have $d_4 > \sqrt{d_3^2 + r_2^2}$. Their workspace admits 2 node points and no voids. Their workspace is not t-connected.

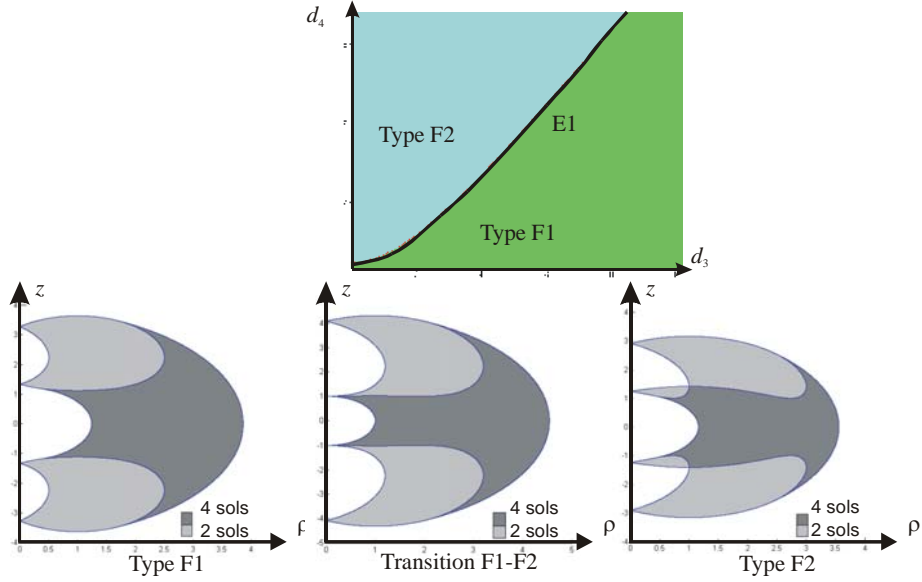

Fig. 9: Parameter space of case F and workspaces of the manipulators having the following parameters: Type F1 ($d_3=2$, $d_4=1.5$, $r_2=1$ and $r_3=1$); Transition F1-F2 ($d_3=2$, $d_4=2.24$, $r_2=1$ and $r_3=1$); Type F2 ($d_3=1$, $d_4=2$, $r_2=1$ and $r_3=1$);

### 3.7 Case G ($d_2=0$, $r_2=0$, $d_3 \neq 0$ and $r_3 \neq 0$)

The manipulators of this case are such that $d_2=0$ and $r_2=0$. Replacing $d_2=0$ and $r_2=0$ in Eq. (10) yields $d_3=0$ or $d_4=0$ or $r_3=0$. Thus, ($\Sigma 2$) must be discarded. On the other hand, Eq. (8) appears to be a spurious solution as we verified numerically that is plays no role. Thus, we have only one type of manipulators, which we call Type G:

- **Type G** (Fig. 10)**:** The manipulators of this type do not admit voids nor node points in their workspace, which is composed of only one 4-solution region. The workspace is t-connected. Thus, these manipulators have a well-connected workspace.

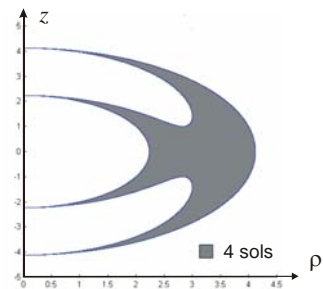

Fig. 10: Workspace of the manipulator having the following parameters
$d_3=1$, $d_4=3$, $r_2=0$ and $r_3=1$.

### 3.8 Case H ($d_2=0$, $r_2 \neq 0$, $d_3=0$ and $r_3 \neq 0$)

This family of manipulators is defined by $d_2=0$ and $d_3=0$. In this case, Eq. (10) is associated with the apparition of nodes on the $z$-axis. Like in the preceding case, Eq. (8) appears to be a spurious solution. Thus, we have only one type of manipulators, which we call Type H:

- **Type H** (Fig. 11)**:** The manipulators of this type do not admit voids nor node points in their workspace, which is composed of only one 4-solution region. The workspace is t-connected. Thus, these manipulators have a well-connected workspace.

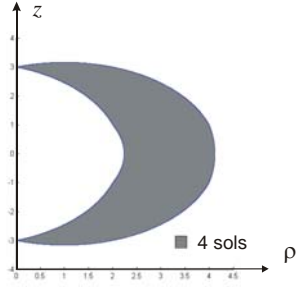

Fig. 11: Workspace of the manipulator having the following parameters $d_3=0$, $d_4=1$, $r_2=3$ and $r_3=1$.

3.9   Case I ($d_2 \neq 0$, $r_2=0$, $d_3 \neq 0$ and $r_3 \neq 0$)

This family of manipulators is defined by $r_2=0$. Like in the preceding two cases, ($\Sigma 1$) appears to be a spurious solution. Replacing $r_2=0$ in Eq. (10) and normalizing by $d_2$ yields:

$$(\Sigma 2): d_4 = \delta \text{ with } \delta = \sqrt{1 + \frac{r_3^2}{d_3^2 - 1}} \qquad (18)$$

The resulting equation depends on three parameters. However, it is possible to study 2-dimensional sections of the parameter space. For a given value of $r_3$, ($\Sigma 2$) defines one or two transition curves depending on $r_3$. When there are two curves, these two curves are located on each side of the asymptote $d_3=1$ where ($\Sigma 2$) is not defined, and which gives rise to an additional transition curve. For this family of manipulators, the maximum number of domains is four and can be shown in a section of the parameter space at a value of $r_3$ where ($\Sigma 2$) defines two curves. Figure 12 shows such a section ($r_3=0.5$) with the four domains, the three transition curves and the workspace of a manipulator in each domain and on each transition.

For this family, there are four types of manipulators, referred to as I1, I2, I3 and I4, respectively:

- **Type I1:** The manipulators of this type have $d_3 > d_2$ and $d_4 > \delta$. Their workspace is free of voids and node points. Their workspace is not t-connected.

- **Type I2:** The manipulators of this type have $d_3 > d_2$ and $d_4 < \delta$. Their workspace admits two node points and one void. Their workspace is not t-connected.

- **Type I3:** The manipulators of this type have $d_3 < d_2$ and $d_4 > \delta$. Their workspace admits one void and no node points. Their workspace is t-connected.

- **Type I4:** The manipulators of this type have $d_3 < d_2$ and $d_4 < \delta$. Their workspace admits two node points and one void. Their workspace is not t-connected.

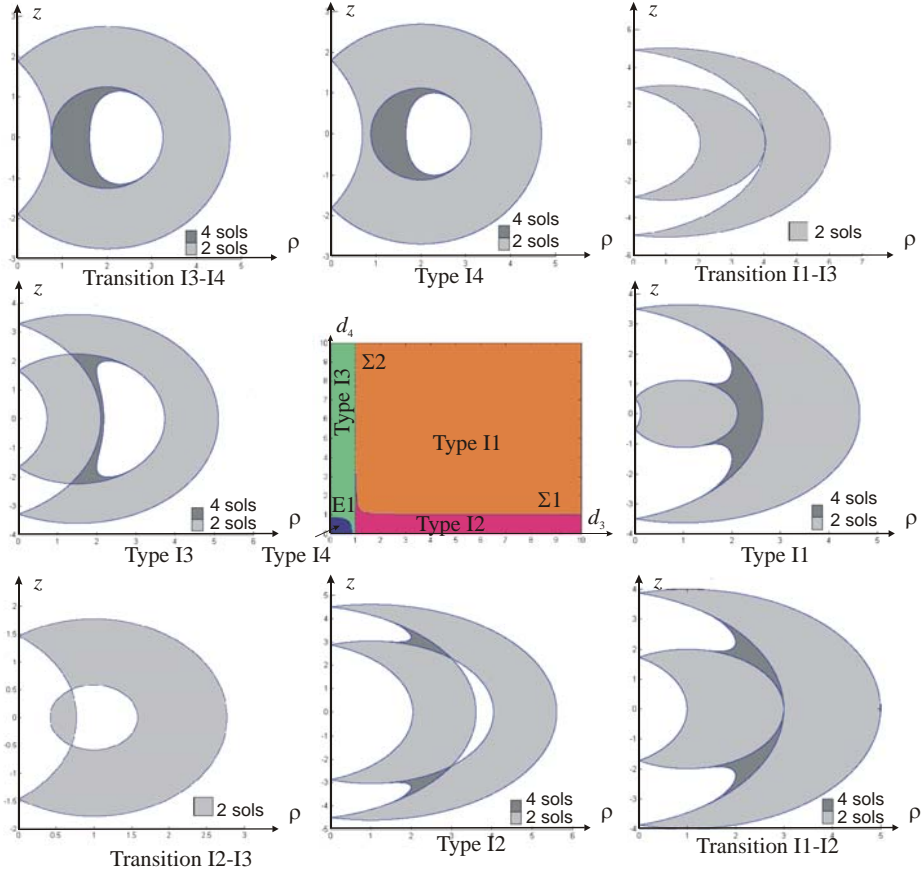

Fig. 12: Parameter space of case I with $r_3=0.5$ and workspaces of the manipulators having the following parameters: Type I1 ($d_2=1$, $d_3=2.5$ and $d_4=1.5$); Type I2 ($d_2=1$, $d_3=3$ and $d_4=0.7$); Type I3 ($d_2=1$, $d_3=0.5$ and $d_4=0.7$); Type I4 ($d_2=1$, $d_3=0.3$ and $d_4=2$); Transition I1-I2 ($d_2=1$, $d_3=3$ and $d_4=1$); Transition I1-I3 ($d_2=1$, $d_3=1$, $d_4=4$); Transition I2-I3 ($d_2=1$, $d_3=1$, $d_4=0.7$); Transition I3-I4 ($d_2=1$, $d_3=0.2$, $d_4=0.8$).

### 3.10 Case J ($d_2 \neq 0$, $r_2=0$, $d_3=0$ and $r_3 \neq 0$)

The manipulators of this case are such that $r_2=0$ and $d_3=0$. Equation (10) is associated with the apparition of nodes on the $z$-axis. On the other hand, Eq. (8) can be shown to be a spurious solution. For this family, thus, there is only one type of manipulators, which we call Type J.

- **Type J** (Fig. 13)**:** The manipulators of this type admit one void and no node points in their workspace, which is composed of only one 4-solution region. Their workspace is t-connected.

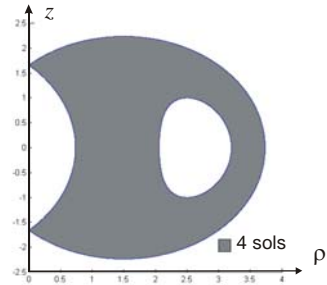

Fig. 13: Workspace of the manipulator having the following parameters $d_2=1$, $d_4=2$, and $r_3=1$.

## 4   Classification synthesis and discussion

We have classifed the ten families of manipulators of Fig. 2 as function of the topology of their workspace. Overall, twenty two types of manipulators have been analyzed. For all manipulators of one given type, the following global kinematic properties are the same: (i) number of nodes (ii) number of voids (iii) number of 2-solution and 4-solutions regions (iv) t-connectivity and well-connectivity of the workspace. The classification of the ten families is synthesized in Tab. 1 below. It is apparent that some manipulator types have better properties than others. Besides, several manipulator types have similar properties. From Tab. 1, we conclude that five manipulator types, namely, types B1, C, E, G and H, have a well-connected workspace (workspace is fully reachable with four inverse kinematic solutions and fully t-connected) (Fig. 14). On the other hand, manipulator types A3, D1, D6, I2, I3 and I4 have poor kinematic performances and should be discarded by the designer. This is an interesting information for the design of alternative manipulator geometries.

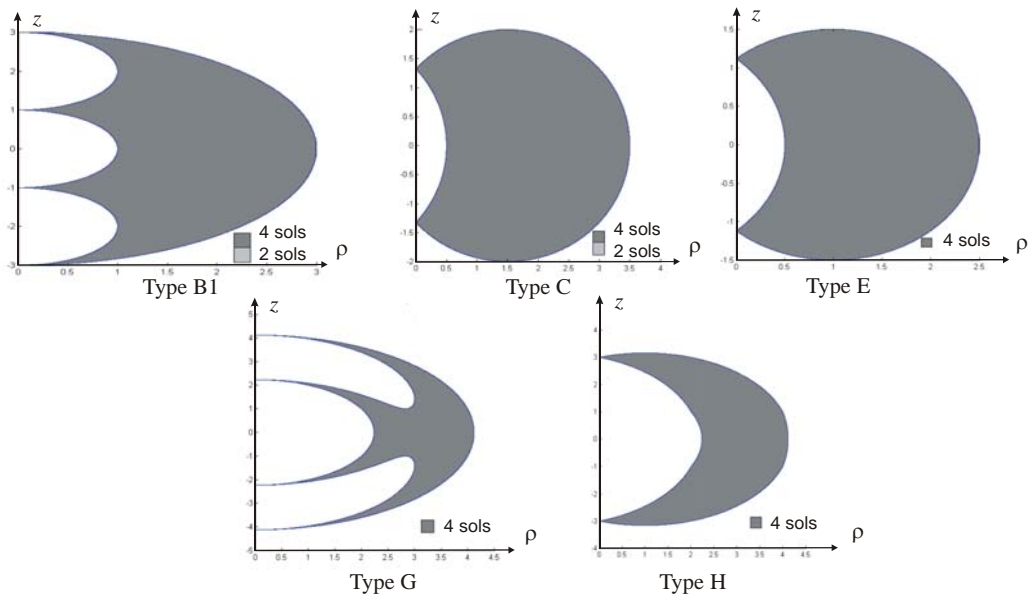

Fig. 14: The five manipulator types with well-connected workspace.

## 5   Conclusions

In this article, the exhaustive classification and enumeration of all types of workspace topology was conducted for the ten families of orthogonal manipulators that have at least one geometric parameter equal to zero. Twenty-two different types of manipulators were identified, which have similar global kinematic properties. The resulting classification has shown that five manipulators type (types B1, C, E, G and H) have a well-connected workspace, which was shown in the past to be specific to Puma-type manipulators. These types are interesting candidates for the design of alternative manipulator geometries.

Finally, it should be recalled that our classification relies only on the topology of the workspace. Other interesting kinematic features, such as the compactness of the workspace or the global conditioning index, were not analyzed. In fact, our classification can be regarded as a preliminary step in the design of new manipulators.

| Type | DH conditions | Void | Node points | 4-solution region | Workspace t-connected | Workspace well-connected |
|---|---|---|---|---|---|---|
| B1 | $d_2=0$, $r_2=0$, $r_3=0$ $d_3>d_4$ | 0 | 0 | All the workspace | Yes | Yes |
| C | $d_3=0$, $r_3=0$ | 0 | 0 | All the workspace | Yes | Yes |
| E | $d_3=0$, $r_2=0$, $r_3=0$ | 0 | 0 | All the workspace | Yes | Yes |
| G | $d_2=0$, $r_2=0$ | 0 | 0 | All the workspace | Yes | Yes |
| H | $d_2=0$, $d_3=0$ | 0 | 0 | All the workspace | Yes | Yes |
| A1 | $d_2=0$, $r_3=0$, $d_4<d_3$ | 0 | 0 | | Yes | No |
| D5 | $r_2=0$, $r_3=0$, $d_3<d_4<d_2$ | 0 | 0 | | Yes | No |
| F1 | $d_2=0$, $d_4 < \sqrt{d_3^2 + r_2^2}$ | 0 | 0 | | Yes | No |
| D2 | $r_2=0$, $r_3=0$, $d_2<d_4<d_3$ | 0 | 0 | | No | No |
| I1 | $r_2=0$, $d_3 > d_2$ and $d_4 > \delta$ | 0 | 0 | | No | No |
| B2 | $d_2=0$, $r_2=0$, $r_3=0$, $d_3<d_4$ | 0 | 1 | All the workspace | No | No |
| D3 | $r_2=0$, $r_3=0$, $d_2<d_3<d_4$ | 0 | 1 | | No | No |
| A2 | $d_2=0$, $r_3=0$, $d_3<d_4<\sqrt{d_3^2+r_2^2}$ | 0 | 2 | | Yes | No |
| D4 | $r_2=0$, $r_3=0$, $d_3<d_2<d_4$ | 0 | 2 | | No | No |
| F2 | $d_2=0$, $d_4 > \sqrt{d_3^2 + r_2^2}$ | 0 | 2 | | No | No |
| A3 | $d_2=0$, $r_3=0$, $d_4 > \sqrt{d_3^2 + r_2^2}$ | 0 | 4 | | Yes | No |
| D6 | $r_2=0$, $r_3=0$, $d_4<d_3<d_2$ | 1 | 0 | Null | Yes | No |
| I3 | $r_2=0$, $d_3 < d_2$ and $d_4 > \delta$ | 1 | 0 | | Yes | No |
| J | $r_2=0$ and $d_3=0$ | 1 | 0 | All the workspace | Yes | No |
| D1 | $r_2=0$, $r_3=0$, $d_4>d_2>d_3$ | 1 | 2 | | No | No |
| I2 | $r_2=0$, $d_3 > d_2$ and $d_4 < \delta$ | 1 | 2 | | No | No |
| I4 | $r_2=0$ | 1 | 2 | | Yes | No |

Tab.1 All types and their kinematic properties.

A useful next step is a detailed inspection of the domains of the parameter space that are associated with the most interesting types, namely, types B1, C, E, G and H. This analysis can be conducted like in [20] on the basis of global kinematic indices such as workspace compactness and global conditioning index. The resulting analysis would make it possible to pinpoint small domains in the parameter space where the manipulators feature, in addition to a well-connected workspace, good workspace compactness or good global conditioning.